%% file: main.tex
\documentclass{Interspeech2024}
\input{preamble}

\interspeechcameraready

\title{Audio Dialogues: \\ Dialogues dataset for audio and music understanding}

\name[]{Arushi}{Goel$^\dagger$}
\name[]{Zhifeng}{Kong}
\name[]{Rafael}{Valle$^\ddagger$}
\name[]{Bryan}{Catanzaro}

\address{NVIDIA}
\email{$^\dagger$arushig@nvidia.com, $^\ddagger$rafaelvalle@nvidia.com}

\keywords{Multi-turn dialogues, Instruction-tuning, AudioLLMs, Audio and Music understanding}

\begin{document}

\maketitle

\begin{abstract}

Existing datasets for audio understanding primarily focus on single-turn interactions (\ie~audio captioning, audio question answering) for describing audio in natural language, thus limiting understanding audio via interactive dialogue. To address this gap, we introduce \texttt{Audio Dialogues}: a multi-turn dialogue dataset containing 163.8k samples for general audio sounds and music. In addition to dialogues, \texttt{Audio Dialogues} also has question-answer pairs to understand and compare multiple input audios together. \texttt{Audio Dialogues} leverages a prompting-based approach and caption annotations from existing datasets to generate multi-turn dialogues using a Large Language Model (LLM). We evaluate existing audio-augmented large language models on our proposed dataset to demonstrate the complexity and applicability of \texttt{Audio Dialogues}.
Our code for generating the dataset will be made publicly available. Detailed prompts and generated dialogues can be found on the demo website\footnote{\url{https://audiodialogues.github.io/}}. 

\end{abstract}

\section{Introduction}
\label{sec.intro}
\input{00intro.tex}

\section{Related work}
\label{sec.related}
\input{01rel.tex}

\section{Data generation pipeline}
\label{sec.data}
\input{02data.tex}

\section{Audio Dialogues dataset}
\label{sec.dialogues}

\input{03dialogues.tex}

\section{Experiments}
\label{sec.exp}
\input{04exp.tex}



\bibliographystyle{IEEEtran}
\bibliography{mybib}

\end{document}

%% file: preamble.tex
\usepackage{times}
\usepackage{microtype}
\usepackage{graphicx}
\usepackage{subfigure}
\usepackage{booktabs} 
\usepackage{lipsum}  

\usepackage{hyperref}

\usepackage{amsmath}
\usepackage{amsfonts}
\usepackage{amssymb}
\usepackage{mathtools}
\usepackage{amsthm}
\usepackage{tikz}
\usepackage[capitalize,noabbrev]{cleveref}

\theoremstyle{plain}

\theoremstyle{definition}

\theoremstyle{remark}

\usepackage[textsize=tiny]{todonotes}

\usepackage{subcaption}
\usepackage{booktabs} 
\usepackage{xcolor} 
\usepackage{url}
\usepackage{pgfplots,pgfplotstable}
\pgfplotsset{compat=1.18}
\pgfplotsset{scaled y ticks=false, scaled x ticks=false}

\definecolor{forestgreen}{rgb}{0.13, 0.55, 0.13}
\definecolor{fulvous}{rgb}{0.86, 0.52, 0.0}
\definecolor{glaucous}{rgb}{0.38, 0.51, 0.71}
\definecolor{lava}{rgb}{0.81, 0.06, 0.13}
\definecolor{buff}{rgb}{0.94, 0.86, 0.51}
\definecolor{chromeyellow}{rgb}{1.0, 0.65, 0.0}
\definecolor{brightube}{rgb}{0.82, 0.62, 0.91}

\hypersetup{
    pdfborderstyle={/S/U/W 1}, 
    linkbordercolor=red,       
    citebordercolor=green,     
    filebordercolor=magenta,   
    urlbordercolor=cyan        
}

\newcommand{\etal}{\textit{et al.}}
\newcommand{\eg}{\textit{e.g.}}
\newcommand{\ie}{\textit{i.e.}}

\usepackage{interval}
\usepackage{tabularx}
\usepackage{color}
\usepackage{xspace}
\usepackage{mathtools}  
\usepackage{siunitx}    
\usepackage{csvsimple}  
\usepackage{placeins}

\usepackage{multirow}
\usepackage{arydshln}
\usepackage[many]{tcolorbox}    	
\usepackage{setspace}               
\usepackage{lipsum,lmodern}
\AtBeginEnvironment{tcolorbox}{\small}

\usepackage[algo2e,ruled,vlined,linesnumbered]{algorithm2e} 
\usepackage{wrapfig}

%% file: 00intro.tex
Audio, a fundamental component of human communication and interaction, carries vast amounts of information, ranging from speech and music to general and ambient sounds. The development of models for audio understanding plays a vital role in various tasks such as audio and sound monitoring \cite{foggia2015audio, alsina2017homesound, chandrakala2019environmental}, speech recognition \cite{chen2023salm, radford2023robust, gong2023whisper}, music recommendation systems \cite{van2013deep, liu2023music, doh2023lp}, and even aiding individuals with hearing impairments. To aid progress in audio understanding, the research community has developed large-scale datasets \cite{doh2023lp, liu2023music, manco2023song, mei2023wavcaps, gemmeke2017audio} (\eg~AudioSet \cite{gemmeke2017audio}, WavCaps \cite{mei2023wavcaps}, AudioCaps \cite{kim2019audiocaps} \textit{etc.}). Hence, models \cite{CLAP2022, deshmukh2023pengi, chu2023qwen, gong2023listen} built on these datasets have shown great potential in learning audio representations for tasks such as audio retrieval \cite{ghosh2023recap, oncescu2021audio}, audio and music captioning \cite{huang2023audiogpt, salewski2023zero, agostinelli2023musiclm, liu2023music}, sound event classification \cite{fonseca2021fsd50k} and so on.

Recent developments have enhanced audio understanding models by integrating them with Large Language Models (LLMs) \cite{touvron2023llama, achiam2023gpt}. This integration has demonstrated the potential to harness the capabilities of LLMs for robust knowledge retention, reasoning, and task execution in audio-related domains \cite{adigwe2018emotional, fonseca2021fsd50k, kim2019audiocaps}. A notable advancement in LLMs is their adeptness to engage in dialogues with humans \cite{peng2023instruction, achiam2023gpt}. It is essential to extend this capability for multi-turn dialogues to audio-augmented LLMs \cite{rubenstein2023audiopalm, peng2023instruction, chu2023qwen, audioflamingo}, as it is pivotal for constructing models capable of listening and interacting effectively. Such advancement in audio models requires conversation-based dialogue datasets tailored for audio applications.

Although there are extensive datasets describing audio or music in natural language \cite{CLAP2023, gardner2023llark}, they are primarily designed for audio captioning \cite{kim2019audiocaps} or single-turn audio question answering \cite{lipping2022clotho, liu2023music}. For instance, these datasets typically include questions such as ``What is the emotion in this audio?" or ``What does the audio convey?". However, fine-tuning models on such datasets limits the potential of audio-augmented Large Language Models (LLMs) to engage in more complex interactions regarding the audio content. Addressing this limitation, Chu~\etal~\cite{chu2023qwen} introduced a multi-turn dialogue dataset consisting of 20,000 samples to train their proposed model. Nonetheless, the dataset's size remains relatively small, and there is a lack of information regarding its generation process.

To address these limitations, in this paper, we propose \texttt{Audio Dialogues}: an audio-based dialogue dataset with \textit{multi-turn dialogues} and \textit{comparison questions} for general sounds and music. Similar in spirit to how instruction-tuning datasets are generated to train vision and language assistants \cite{yang2023re, liu2023visual, alayrac2022flamingo}, we use prompting-based approach to generate a multi-turn dialogue dataset for audios using a pre-trained LLM \cite{achiam2023gpt, touvron2023llama}. Specifically, we utilize caption annotations sourced from the AudioSet strongly labeled dataset \cite{gemmeke2017audio} and the MusicCaps dataset \cite{agostinelli2023musiclm} to guide the dialogue generation process leveraging GPT-4 \cite{achiam2023gpt}. Additionally, we implement a data filtration strategy to filter out noisy synthetic dialogues, promoting the retention of the most reliable ones. In total, our proposed dataset comprises of 163.8k samples, each containing between one to four dialogues. 

Our main contributions are as follows: 1) a multi-turn dialogue dataset, \texttt{Audio Dialogues} for general sounds and music understanding with training and evaluation splits, 2) a detailed data generation pipeline to foster the generation of dialogue datasets, and 3) evaluation of existing audio-augmented large language models \cite{chu2023qwen, audioflamingo, gong2023ltu} on our proposed dataset.

%% file: 01rel.tex
\textbf{Instruction tuning datasets. }
Instruction following Large Language models (LLMs) \cite{touvron2023llama, achiam2023gpt} have shown remarkable capabilities in zero-shot and few-shot tasks in the language domain such as machine translation \cite{wang2022progress}, summarization \cite{el2021automatic} and so on. This idea of developing models that can follow instructions has then been extended to other domains such as vision \cite{yang2023re, alayrac2022flamingo} and audio \cite{huang2023audiogpt, audioflamingo, peng2023instruction}. LLaVA \cite{liu2023visual} made the first attempt at generating instruction-following data involving visual content using GPT-4. Specifically, they use image captions and bounding box localization as meta information for the image to be used as query for the language model. Overall, they collect 158k samples for language-image instruction-following data. Since then, there has been growing interest in developing instruction following datasets such as VALLEY \cite{luo2023valley}, Macaw-LLM \cite{lyu2023macaw} and Video-ChatGPT \cite{maaz2023video}. 

In the audio domain, LTU \cite{gong2023listen} generated, using GPT, an open-ended question-answering dataset that tries to capture general knowledge and reasoning ability about general sounds. LTU's audio based question answering dataset is limited, given that it only has single-turn conversations, lacks in complex context between conversations and does not have strong correlations between rounds (\eg~use of pronouns). Qwen-Audio \cite{chu2023qwen} curates a 20k audio-based instruction-following dataset, but there is little to no discussion about the curation process and dataset. Our \texttt{Audio Dialogues} dataset addresses all the above mentioned limitations by generating multi-turn conversations for an audio sample, covering both general sounds and music domains. Compared to existing datasets, \texttt{Audio Dialogues} has multi-turn dialogues with strong correlations between rounds through the presence of pronouns (\eg~he, she, it), follow-up questions based on the previous answer, and complex context.
\vspace{-0.03cm}

\noindent
\textbf{Audio augmented LLMs}. 
Recent research has focused on advancing audio foundation models \cite{rubenstein2023audiopalm, huang2023audiogpt, deshmukh2023pengi, audioflamingo} capable of comprehending audio content by harnessing Large Language Models (LLMs) \cite{touvron2023llama, achiam2023gpt}. Typically, these models employ an audio encoder \cite{CLAP2023, gong2023whisper} to convert audio into tokens, which are then integrated with textual instructions within an LLM to generate the final response. These models are pretrained on various tasks such as audio captioning \cite{kim2019audiocaps, mei2023wavcaps}, emotion recognition \cite{cao2014crema, busso2008iemocap, barros2018omg}, sound event classification \cite{fonseca2021fsd50k}, speech recognition \cite{radford2023robust, gong2023whisper}, music understanding \cite{agostinelli2023musiclm, musdb18-hq, liu2023music}, among others, and have shown significant gains in zero-shot and few-shot performances using a unified model. While these models exhibit robust audio comprehension, recent works like Audio Flamingo \cite{audioflamingo} have introduced techniques such as in-context learning \cite{alayrac2022flamingo} and retrieval-augmented generation \cite{yang2023re} to enhance the model's instruction-following capabilities through fine-tuning with interleaved audio-text pairs. To measure the importance of our proposed \texttt{Audio Dialogues} dataset, we evaluate the performance of audio foundation models such as LTU \cite{gong2023listen}, Qwen-Audio \cite{chu2023qwen} and Audio Flamingo \cite{audioflamingo} on these multi-turn dialogues.

%% file: 02data.tex
\begin{figure*}[!ht]
    \centering
    \includegraphics[width=0.7\linewidth]{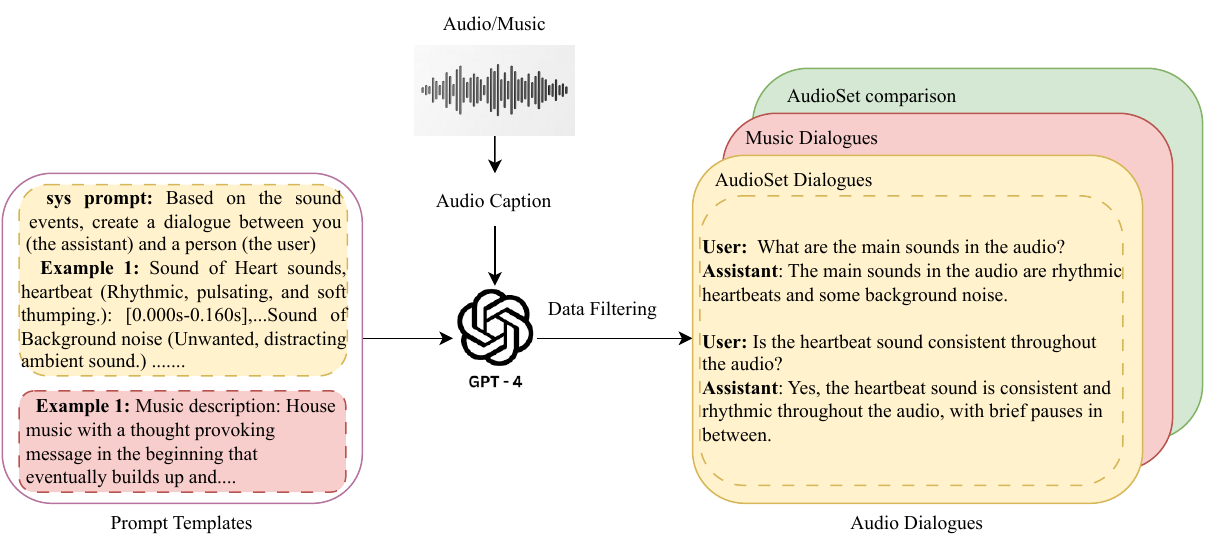}
    \vspace{-0.5em}
    \caption{Illustration of our data generation pipeline. \texttt{Audio Dialogues} is generated using GPT-4 which takes text-only inputs to generate subsets AudioSet dialogues, Music dialogues and AudioSet comparison subsets of our proposed dataset.}
    \label{fig.datapipeline}
\end{figure*}

\subsection{Pipeline}
\label{sec.pipleine}
In this section, we discuss our data generation pipeline illustrated in \cref{fig.datapipeline}. 
We construct \texttt{Audio Dialogues} using the strongly labeled AudioSet-SL \cite{hershey2021benefit} and MusicCaps \cite{agostinelli2023musiclm}.
The Audioset-SL dataset \cite{hershey2021benefit} has time-stamped annotations for the 10-second audio clips which we pre-process to describe sound events for the audio samples.

Following \cite{gong2023listen}, we augment timestamped sound events with acoustic features for each sound event. Specifically, given a sound class name, \eg~howl, we prompt GPT-4 with ``describe the acoustic characteristic of a \textit{howl} sound in less than 10 words.'' This gives us an acoustic feature description of that sound class. An augmented sound event with acoustic features example is shown below:  

\begin{table}[!h]
    \centering
    \begin{tabular}{|p{7.5cm}|}
        \hline
\textbf{Sound events:}
Sound of Howl (Loud, prolonged, mournful, echoing sound.): [0.406s-9.237s], [9.575s-10.000s]; Sound of Wind noise (microphone) (Low frequency, random, broadband sound.): [2.128s-2.584s], [9.288s-9.850s]; Sound of Animal (Loud, diverse, and often rhythmic.): [8.174s-9.221s], [9.778s-10.000s].\\
\hline
\end{tabular}
\end{table}

Similar to AudioSet-SL, the MusicCaps dataset \cite{agostinelli2023musiclm} has detailed descriptions or captions for the music samples which we directly use as input for dialogue generation.

We utilize these audio sound events and music description information along with prompt templates to guide GPT-4 \cite{achiam2023gpt} in generating multi-turn dialogues. Next, we discuss the prompt design and propose a data filtration strategy to retain only high quality dialogues.

\subsection{Prompts}
\label{sec.dialogue_prompts}
We design specific prompt templates to generate 1) multi-turn dialogues and, 2) audio comparison question-answer pairs. 
To generate multi-turn dialogues for the Audioset-SL \cite{hershey2021benefit} and MusicCaps \cite{agostinelli2023musiclm} dataset, the prompt template consists of a \textit{system prompt} and \textit{examples} of hand-crafted dialogues. Due to space constraints, we only show an example of system prompt to generate the subset AudioSet Dialogues and Music Dialogues. Detailed prompts are on our demo website.

\begin{table}[!h]
    \centering
    \begin{tabular}{|p{7.5cm}|}
        \hline
\textbf{System prompt}\\
Based on the sound events, create a dialogue between you (the assistant) and a person (the user) about the events in the audio. Each dialogue should consist of:\\
1. A user examines the audio and sends a reasonable and creative message to the assistant. \\
2. Once the audio is provided, the assistant thoroughly perceives and comprehends them, responding with helpful answers that provide comprehensive reasoning. Do not include timestamps in the answer provided by the assistant.\\
3. Considering the past dialogue i.e. the question and the answer in the previous timestep, the user chooses another question or further inquiry.\\
4. The assistant provides a helpful answer providing comprehensive reasoning regarding the audio content.
\\
All the questions asked by the user should be diverse and complex.\\
\hline
\end{tabular}
\end{table}

Apart from the multi-turn dialogues, the \texttt{Audio Dialogues} dataset also consists of question-answer pairs for multiple (two or more) audios, called as AudioSet comparison subset. Below we show the prompt template used to generate these question-answer pairs given the meta information from the audio clips. 
\begin{table}[!h]
    \centering
    \begin{tabular}{|p{7.5cm}|}
        \hline
\textbf{System prompt}\\
Based on the description of audios, create a dialogue between you (the assistant) and a person (the user) about the events in the audio. \\
\textbf{Example 1:}
\\
Audio 1: Sound of Car (Engine hum and tire noise.)\\
Audio 2: Sound of Car (Engine hum and tire noise.)
\\
User: What's the common type of sound in these two audios? \\
Assistant: Both of them have sounds of car and engine humming.\\
\hline
\end{tabular}
\end{table}

In order to cluster multiple audios to be used as meta information for generating question-answer pairs, we find similar and dissimilar neighbours (2 or more) for the query audio sample using CLAP~\cite{CLAP2022} embedding features of the audio clips. Specifically, we randomly sample k between 2 and 4 to select top-k or bottom-k audios given CLAP cosine similarity scores.

\subsection{Data filtration}
\label{sec.data_filtration}

Our data generation pipeline in \cref{sec.dialogue_prompts} generates answers that have phrases such as ``difficult to infer", ``not specified", ``no specific", ``no information", and so on. To ensure desirable outputs from the model, we filter these QA pairs following~\cite{gardner2023llark}. This is done by manually designing a list of phrases denoting low confidence or uncertainty in the generated answer.

\begin{figure}[!h]
    \centering
    \includegraphics[width=0.6\linewidth]{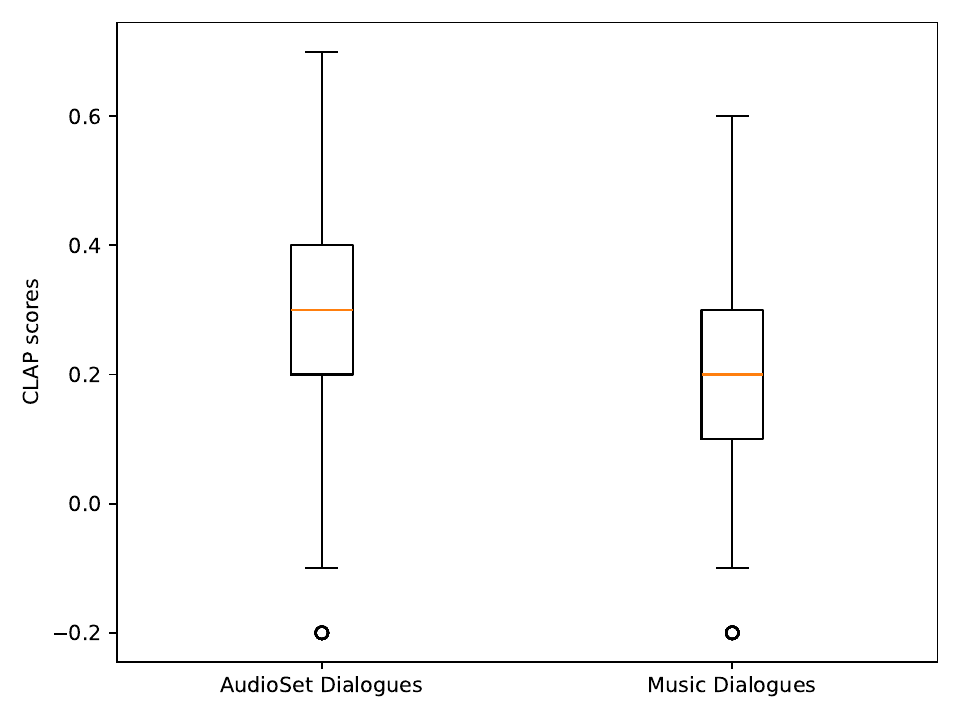}
    \caption{LAION-CLAP similarities before filtration for AudioSet Dialogues (left) and Music Dialogues (right).}
    \label{fig:clap_sim}
\end{figure}

In order to further improve data quality especially the accuracy of the generated answer to the audio, we additionally pick samples in which the answer is highly relevant to the query audio sample. This is done by computing the cosine similarity between the CLAP text-embeddings and audio-embeddings \cite{wu2023large} for each QA pair in each dialogue. In Figure~\ref{fig:clap_sim}, we plot the distributions of these similarities before filtration. The samples that have a similarity of less than 0.3 are removed from the dataset.

%% file: 03dialogues.tex
\noindent
\textbf{Statistics. } 
In \cref{tab:statistics}, we show dataset statistics of Qwen Audio \cite{chu2023qwen} and its instruction fine-tuning dataset for dialogues with 20,000 samples. Apart from this, there is no other information provided as the dataset is not open-source. We also compare the different subsets of our \texttt{Audio Dialogues} dataset. 
For the AudioSet Dialogues subset, there are
76,642 dialogues in the train split and 1,442 dialogues in test split. The Music Dialogues subset has 3,358 dialogues in the train split and 1,641 dialogues in the test split. Each sample has one to four rounds (QA pairs) per dialogue and 1 audio sample as input. 
Our AudioSet Comparison subset has 64,085 dialogues in the train split and 16,249 dialogues in the test split. Compared to all the other subsets, the AudioSet Comparison dataset has 2 to 4 audio samples as input per dialogue, with an average of 3 audios in the entire subset. 
\vspace{-0.3cm}

\begin{table}[!h]
\renewcommand*{\arraystretch}{1.0}
    \resizebox{\linewidth}{!}{
        \begin{tabular}{lccc}
        \toprule
        Dataset & Avg. turn or audio & \multicolumn{2}{c}{\#Samples} \\
        & per dialogue & Train & Test \\ 
        \toprule
        Qwen-Audio~\cite{chu2023qwen} & -/1 & $\sim$20,000 & N/A \\
        \midrule
        \textbf{Audio Dialogues (Ours)} &  & & \\
        AudioSet Dialogues & 2.21/1 & 76,642 & 1,442  \\
        Music Dialogues & 1.67/1 & 3,358 & 1,641 \\
        AudioSet Comparison & 1/3.00  & 64,085 & 16,249\\
        \bottomrule
    \end{tabular}}
    
    \centering
    \caption{Average number of turns or average number of audios per dialogue, and total number of train and test samples for all the subsets of \texttt{Audio Dialogues}. }
    \vspace{-0.5cm}

    \label{tab:statistics}
\end{table}

\vspace{-0.3cm}
\noindent
\textbf{Generated dialogues.} We demonstrate an example of generated dialogue on the Audioset-SL \cite{hershey2021benefit} dataset in \cref{tab:gen_dial}. 
The generated dialogue captures various aspects of the audio by capturing the frequency of events in the audio, their variation and impact, and the ordering of the events.
More examples of all the other subsets are on the demo website.

%% file: 04exp.tex
In this section, we evaluate three recent audio understanding LLMs on \texttt{Audio Dialogues}: LTU \cite{gong2023ltu}, Qwen-Audio \cite{chu2023qwen} and Audio Flamingo \cite{audioflamingo}. We used the pretrained LTU and Qwen-Audio checkpoints and re-implemented Audio Flamingo. The results are in \cref{tab: multi-turn results}.\footnote{We do not report Qwen-Audio results on Music Dialogues as it was not trained on music understanding datasets.} We first do zero-shot evaluation.

\begin{table}[!h]
    \centering
    \begin{tabular}{|p{7.5cm}|}
        \hline
\texttt{[Audio annotations: Sound of Wind (Whistling, rustling, airy sound.): [0.000s-10.000s]; Sound of Change ringing (campanology) (Melodic, rhythmic bell sequences.): [0.116s-10.000s]; Sound of Hubbub, speech noise, speech babble (Loud, overlapping, indistinct chatter.): [0.116s-10.000s]; Sound of Tap (Sharp, rhythmic, resonant sound.): [0.465s-0.693s], [2.394s-2.543s]; Sound of Clapping (Sharp, brief, percussive sound.): [7.551s-7.693s]]}

~~

User: What are the main sounds in the audio?\\
Assistant: The main sounds in the audio are wind, change ringing or campanology, hubbub or speech noise, tapping, and clapping.
\\
\\
User: What is the nature of the speech noise?\\
Assistant: The speech noise is loud, overlapping, and indistinct, which is often referred to as hubbub or speech babble.
\\
\\
User: Is there any specific event that the clapping sound might be associated with?\\
Assistant: Given the presence of change ringing and hubbub, the clapping sound could be associated with a celebration or event, possibly in a public place like a town square or church.\\
\hline
\end{tabular}
\caption{Example dialogue from \texttt{Audio Dialogues}.}
\label{tab:gen_dial}
\end{table}
%
\vspace{-0.6cm}
We then fine-tuned LTU and Audio Flamingo on the training set of AudioSet Dialogues and Music Dialogues. For the AudioSet comparison subset, we only fine-tune Audio Flamingo as LTU does not support multiple audios as input.  Compared to the zero-shot results, the fine-tuned model achieves better performance across all metrics (marked as $\Delta$). As Audio Flamingo is trained with retrieval and in-context learning, it shows better performance and can use context better than LTU. This shows fine-tuning on \texttt{Audio Dialogues} enables an audio understanding LLM to have much stronger dialogue capabilities.

\begin{table}[!t]
    \renewcommand*{\arraystretch}{1.0}

    \resizebox{\linewidth}{!}{
    \begin{tabular}{llccc}
    \toprule
    Subset & Method & CIDEr $\uparrow$ & Bleu4 $\uparrow$ & R-L $\uparrow$ \\ \toprule 
    \multirow{5}{*}{AudioSet Dialogues}
    & Qwen-Audio & $0.507$ & $0.060$ & $0.292$ \\
    & LTU & $0.580$ & $0.122$ & $0.324$ \\
    & LTU$^\Delta$ & $0.823$ & $0.153$ & $0.403$  \\
    & Audio Flamingo & $0.615$ & $0.072$ & $0.341$ \\
    & Audio Flamingo$^\Delta$ & $\mathbf{1.672}$ & $\mathbf{0.244}$ & $\mathbf{0.477}$ \\ \midrule
    \multirow{4}{*}{Music Dialogues}
    & LTU & $0.168$ & $0.065$ & $0.217$ \\
    & LTU$^\Delta$ & $0.419$ & $0.108$ & $0.336$ \\
    & Audio Flamingo & $0.395$ & $0.036$ & $0.250$ \\
    & Audio Flamingo$^\Delta$ & $\mathbf{1.191}$ & $\mathbf{0.155}$ & $\mathbf{0.426}$ \\ \midrule
    AudioSet Comparison & Audio Flamingo$^\Delta$ & $1.885$ & $0.444$ & $0.627$ \\ 
    \bottomrule
    \end{tabular}}
    \centering
    \caption{Evaluation of LTU \cite{gong2023listen}, Qwen-Audio \cite{chu2023qwen} and Audio Flamingo \cite{audioflamingo} on the subsets of \texttt{Audio Dialogues} test sets.  We report the following metrics: CIDEr \cite{vedantam2015cider}, Bleu4 \cite{papineni2002bleu}, and Rouge-L (R-L) \cite{lin2004rouge}. Scores improve for all models fine-tuned on \texttt{Audio Dialogues}, marked $\Delta$.}
    \vspace{-2em}
    \label{tab: multi-turn results}
\end{table}
In \cref{tab:qual_results}, we also present qualitative results on the test set of the \texttt{Audio Dialogues} dataset. We find Audio Flamingo generates more accurate answers than LTU in the multi-turn dialogue setting.

\section{Discussion}
In this paper, we introduce the Audio Dialogues dataset, designed to enhance audio understanding within the context of multi-turn dialogues, covering a broad spectrum of general sounds and music. By leveraging a prompting-based approach and utilizing caption annotations from existing datasets, we generate a substantial volume of high-quality dialogues suitable for training and evaluating audio-based models.

While \texttt{Audio Dialogues} enables models to engage in interactions with the user, leading to better comprehension and more nuanced responses, there are certain limitations. One limitation is the absence of timestamps in the generated questions and answers. In the future, an interesting direction would be to temporally ground an answer about an audio. Another limitation is the need for strong annotations to generate dialogues from LLMs. A potential direction to scale generated data is to generate detailed meta descriptions from an existing audio understanding model and designing human feedback-based data filtration strategies. 
\vspace{-0.3cm}
\begin{table}[!h]
    \centering
    \begin{tabular}{|p{7.8cm}|}
        \hline

\texttt{[Audio annotations: Sound of Rattle (Sharp, rapid, repetitive, percussive sound.): [0.378s-1.346s]; Sound of Spray (High-pitched, hissing, and continuous sound.): [1.402s-1.921s], [2.024s-4.346s]]}

\textbf{User}: \textit{Is the rattle sound followed immediately by the spray sound?}\\
\textbf{GT}: \textit{Yes, the rattle sound is immediately followed by the spray sound with a very short gap in between.}

\textbf{LTU}: \textit{Yes, the rattle sound is followed immediately by the spray sound. They occur in quick succession.}

\textbf{Audio Flamingo}: \textit{Yes, the rattle sound is followed immediately by the spray sound.}

~~

\textbf{User}: \textit{Does the spray sound occur only once?}

\textbf{GT}: \textit{The spray sound occurs twice in the audio.}

\textbf{LTU}: \textit{Yes, the spray sound occurs only once in the audio, between the 1.414 second and 2.036 second marks. It's a high-pitched, rapid, and hissing sound.}

\textbf{Audio Flamingo}: \textit{No, the spray sound occurs twice in the audio.}

~~

\textbf{User}: \textit{ Is the second spray sound longer than the first?}

\textbf{GT}: \textit{Yes, the second spray sound lasts longer than the first one.}

\textbf{LTU}: \textit{Yes, the second spray sound is longer than the first one.}

\textbf{Audio Flamingo}: \textit{Yes, the second spray sound is longer than the first one.}\\
\hline
\end{tabular}

\caption{Qualitative results on \texttt{Audio Dialogues} dataset.}
\label{tab:qual_results}
\end{table}